\documentclass[times,preprint,10pt]{elsarticle}
\usepackage[
  letterpaper,
  top=4.3cm,
  right=4.8cm,
  bottom=4.3cm,
  left=4.8cm
]{geometry}
\usepackage[ruled,vlined,linesnumbered]{algorithm2e}

\SetKwInOut{Input}{Input}
\SetKwInOut{Output}{Output}

\SetNlSkip{0.3em}

\SetAlgoNlRelativeSize{-1}

\SetAlFnt{\fontsize{10pt}{11.5pt}\selectfont}
\SetAlCapFnt{\fontsize{8pt}{9.5pt}\selectfont}
\SetAlCapNameFnt{\fontsize{8pt}{9.5pt}\selectfont}

\DontPrintSemicolon

\SetAlgoLongEnd

\SetKwFor{For}{for}{}{end for}
\SetKwFor{While}{while}{}{end while}
\SetKwIF{If}{ElseIf}{Else}{if}{}{else if}{else}{end if}

\usepackage{booktabs}
\usepackage{caption}
\usepackage{multirow}
\usepackage{array}
\usepackage{placeins}
\DeclareCaptionFont{eightpt}{\fontsize{8pt}{9.5pt}\selectfont}
\DeclareCaptionFont{eightpt}{\fontsize{8pt}{9.5pt}\selectfont}

\usepackage{caption}
\usepackage{amssymb}
\usepackage{amsmath}
\usepackage{float}
\usepackage{multirow}
\usepackage{url}
\journal{Pattern Recognition}

\begin{document}

\begin{frontmatter}



\title{SDM-Q: Cost-Aware Staged Decision-Making for Multi-Omics Classification with Deep Q-Learning}


\author[1]{Nan Mu\fnref{equal}}
\author[1]{Yangfan Xiao\fnref{equal}}
\author[1]{Ling Wang}
\author[1]{Xiaoning Li}
\author[2]{Yue Kang}
\author[3]{Chen Zhao\corref{cor1}}

\address[1]{College of Computer Science, Sichuan Normal University, Chengdu, Sichuan 610101, China.}
\address[2]{Department of Mathematics, College of Science and Mathematics, Kennesaw State University, Kennesaw, GA 30144, USA.}
\address[3]{Department of Computer Science, College of Computing and Software Engineering, Kennesaw State University, Marietta, GA 30060, USA.}

\fntext[equal]{N. Mu and Y. Xiao contributed equally to this work.}

\cortext[cor1]{Corresponding author. Email: czhao@kennesaw.edu (C. Zhao).}

%% Abstract
\begin{abstract}

Multi-omics data provide complementary molecular characterizations of disease phenotypes and play an important role in disease diagnosis and subtype classification in precision medicine. However, acquiring complete multi-omics profiles is expensive and time-consuming, while most existing deep learning methods assume full modality availability during inference, resulting in substantial redundancy and limited practicality in clinical settings. To address this issue, we propose SDM-Q, a reinforcement learning framework for adaptive and cost-aware multi-omics classification. Specifically, multi-omics diagnosis is reformulated as a finite-horizon sequential decision problem, where the currently acquired omics modalities define the diagnostic state at each stage. An action--value function determines whether to acquire an additional modality or terminate the decision process and output the final prediction. To balance diagnostic utility and acquisition cost, the reward is defined only at the terminal stage and jointly determined by classification correctness and cumulative modality acquisition cost. A backward stage-wise optimization strategy is introduced to improve policy consistency and training stability. Experiments on four public multi-omics datasets, including ROSMAP, LGG, BRCA, and KIPAN, demonstrate that SDM-Q effectively reduces redundant modality acquisition while maintaining competitive classification performance compared with methods using complete multi-omics inputs. In the BRCA and KIPAN datasets, more than 99\% and 95\% of subjects, respectively, achieve accurate classification using only a single omics modality, while the average number of acquired modalities remains below two for ROSMAP and LGG. These results suggest that cost-aware sequential decision-making provides an effective paradigm for improving the efficiency of precision medicine workflows.

\end{abstract}


\end{frontmatter}

\section{Introduction}

With the rapid development of precision medicine, multi-omics data integration has become an important paradigm for characterizing molecular heterogeneity and improving disease diagnosis ~\cite{ref1}. Different omics modalities, including gene expression, DNA copy number variation, and DNA methylation, provide complementary biological information from distinct molecular perspectives. By jointly modeling these heterogeneous data sources, multimodal learning frameworks have achieved promising performance in tasks such as tumor subtype classification and prognostic prediction~\cite{ref2}.

Despite these advances, substantial differences remain among omics assays in terms of acquisition procedures, turnaround time, and economic cost, making routine collection of complete multi-omics profiles impractical in real-world clinical settings. In practice, multi-omics diagnosis is inherently a sequential information acquisition process, where subsequent tests are determined according to currently available evidence~\cite{ref33}. However, most existing intelligent diagnostic models~\cite{ref3} formulate multi-omics analysis as a static classification problem that assumes one-shot access to all modalities during inference. This paradigm overlooks the fact that reliable diagnosis can be achieved using only a subset of omics modalities, whereas incorporating additional omics data can produce redundant or marginally informative features, thus increasing cost and causing delay in diagnosis.

More importantly, existing studies~\cite{ref4,ref5,ref6} model acquisition cost from an inappropriate perspective. In practical multi-omics diagnosis, the major resource consumption arises from modality-level laboratory assays, where different omics platforms exhibit substantial heterogeneity in both monetary cost and acquisition latency. Although several studies have explored cost-sensitive learning~\cite{ref4} or reinforcement learning--based adaptive acquisition strategies~\cite{ref5,ref6}, most existing approaches define cost in terms of computational complexity~\cite{ref35} or feature-level acquisition~\cite{ref34}. Consequently, they fail to accurately characterize the modality-level testing cost in clinical workflows, limiting their ability to determine whether additional omics modalities should be acquired for individual subjects.

To address these challenges, we reformulate multi-omics data classification as a subject-specific sequential decision-making problem and propose a \textbf{S}taged \textbf{D}ecision-\textbf{M}aking model based on deep \textbf{Q}-learning \textbf{(SDM-Q)}. Specifically, the diagnostic process is decomposed into multiple decision stages, where each decision is conditioned on the omics modalities acquired thus far. At each stage, SDM-Q determines whether to acquire an additional modality or to terminate the process and produce a classification prediction. By explicitly balancing diagnostic reliability with cumulative acquisition costs, SDM-Q derives individualized and cost-effective testing pathways for different subjects. This design enables on-demand modality acquisition and adaptive diagnosis, making multi-omics learning more compatible with practical clinical workflows.The source code of SDM-Q is publicly available at \url{https://github.com/chenzhao2023/SDM-Q}.

The contributions of this paper are summarized as follows:

1) We propose SDM-Q, which reformulates multi-omics data classification from a static multi-modality classification task into a finite-horizon sequential decision problem. This formulation enables adaptive and subject-specific modality acquisition, allowing the diagnostic process to dynamically determine both the order and the number of required omics modalities to be collected.

2) We develop a unified multi-stage decision mechanism that models modality acquisition and stopping decisions within a Markov Decision Process (MDP). By encoding the currently acquired omics layers into state representations and learning action values for both modality acquisition and classification termination, SDM-Q explicitly captures the trade-off between classification benefit and cumulative classification cost. A cross-stage value propagation strategy is further introduced to improve policy consistency and stability across decision stages.

3) Extensive experiments on four public multi-omics datasets (ROSMAP, LGG, BRCA, and KIPAN) demonstrate that SDM-Q significantly reduces redundant modality acquisition while maintaining classification performance comparable to state-of-the-art methods that rely on complete multi-omics data. These results highlight the effectiveness and practical advantages of staged sequential classification for cost-aware precision medicine.

\section{Related Work}

This section reviews representative studies closely related to our work, with a particular focus on modeling paradigms for multi omics information acquisition and cost-sensitive adaptive inference.

\subsection{Modeling Paradigms for Multimodal Information Acquisition}

In recent years, multimodal learning has become an important paradigm for integrating heterogeneous biomedical data to better characterize complex disease phenotypes. Under this paradigm, multimodal deep learning methods have been widely applied to multi-omics data integration, achieving notable progress. Chaudhary \textit{et al.}~\cite{ref7} developed a deep neural network--based fusion framework that integrates heterogeneous omics data within a shared representation space and demonstrated its effectiveness in liver cancer survival prediction. Wang \textit{et al.}~\cite{ref8} subsequently proposed MOGONET, which employs graph convolutional networks (GCNs) to construct modality-specific graphs and perform feature-level fusion, leading to improved patient classification and biomarker identification. Building on this idea, Tanvir \textit{et al.}~\cite{ref9} introduced MoGAT, which leverages graph attention mechanisms to strengthen cross-modality feature interactions and adaptively reweight different omics modalities for more effective integration.

Despite their success, these approaches follow a static fusion paradigm, where all modalities are assumed to be available before model inference. Such an assumption differs fundamentally from real-world clinical diagnostic workflows. In clinical practice, physicians typically determine the necessity of additional tests based on currently available diagnostic evidence. To alleviate this limitation, Mu \textit{et al.}~\cite{ref10} proposed an uncertainty-aware progressive integration framework, in which additional modalities are introduced sequentially based on predictive uncertainty under a predefined acquisition order. Although this approach introduces a notion of staged information acquisition, its reliance on a fixed modality sequence limits the flexibility to select the most informative modality dynamically at each stage.

In contrast, unlike methods constrained by predefined modality orders, SDM-Q models the diagnostic process as a MDP, where the model directly estimates the action values (Q-values) of acquiring specific candidate modality given the current avaiable modalities. As a result, SDM-Q enables the planning of individualized testing strategies that balance classification accuracy with acquisition cost, thereby improving resource efficiency while maintaining reliable classification results.

\subsection{Cost-Sensitive Learning and Adaptive Inference}

Cost-sensitive learning aims to explicitly balance predictive performance and decision cost, enabling models to make rational inferences under limited resource budgets. In multi-omics data classification, such modeling is particularly important for capturing the trade-off between classification accuracy and cost. However, in most existing studies, cost is defined primarily in terms of computational resource consumption rather than the real-world expenses associated with data acquisition.

Representative works include BranchyNet, proposed by Teerapittayanon \textit{et al.}~\cite{ref4}, which introduces early-exit branches at intermediate network layers to reduce average inference computation. Similarly, the BlockDrop proposed by Wu \textit{et al.}~\cite{ref11}, dynamically selects residual blocks during inference to lower computational complexity. While these methods effectively reduce inference-time computation, the cost they optimize corresponds to computational effort. This notion of cost differs fundamentally from modality acquisition cost in multi-omics medical scenarios, where expenses arise from laboratory assays and sequencing procedures. As a result, the optimization objectives and decision constraints in these settings are inherently different.

To address sequential information acquisition under resource constraints, reinforcement learning (RL) has been introduced into cost-aware prediction tasks. In these approaches, models perform stepwise decisions that balance predictive performance against information acquisition cost. Contardo \textit{et al.}~\cite{ref5} formulated feature acquisition as a sequential decision-making problem, where the model determines whether additional features should be obtained based on the information already observed. Building on this framework, An \textit{et al.}~\cite{ref6} incorporated RL strategies to learn adaptive feature acquisition policies under explicit budget constraints, leading to improved predictive performance. These studies demonstrate the effectiveness of RL for progressive information acquisition and efficient decision-making.

Nevertheless, most RL--based approaches treat individual features or feature subsets as the fundamental decision units, restricting the decision granularity to the feature level. This assumption is not well suited to multi-omics medical settings, where omics data are typically obtained at the modality level through laboratory assays or sequencing experiments. Each acquisition decision therefore introduces an entire set of features simultaneously. Moreover, feature-level decision modeling fails to capture the substantial heterogeneity among omics modalities in terms of acquisition cost, turnaround time, and clinical feasibility, resulting in policies that deviate from real-world diagnostic workflows.

In contrast, this work revisits cost-sensitive modeling from the perspective of modality-level information acquisition. We formulate multi-omics diagnosis as a subject-specific sequential decision-making problem, where omics modalities serve as the fundamental decision units. SDM-Q provides a more practical solution for dynamic information acquisition scenarios characterized by heterogeneous testing costs and highly individualized diagnostic requirements in multi-omics medicine.

\section{Methodology}

Existing multi-omics data classification methods are largely based on a static fusion paradigm, which assumes that all omics modalities are available at the inference stage. However, this assumption overlooks two important characteristics of real-world clinical workflows: the sequential nature of modality acquisition and the substantial heterogeneity in acquisition costs across different omics assays. Consequently, static approaches struggle to effectively balance three competing objectives, i.e., selecting informative modalities, achieving accurate diagnosis, and controlling detection cost.

To address these limitations, we propose SDM-Q, a dynamic decision framework that reformulates multi-omics data classification as a finite-horizon MDP. SDM-Q models the diagnostic process as a sequence of decision stages. At each stage, SDM-Q evaluates the trade-off between the expected diagnostic benefit of candidate modalities and their corresponding acquisition costs based on the currently observed patient state, thereby generating adaptive and subject-specific modality acquisition paths.

However, sequential decision-making in this setting naturally leads to sparse reward signals, since the diagnostic outcome becomes available only at the terminal stage. To enable effective policy learning, we introduce a backward multi-stage training mechanism that propagates terminal classification rewards to earlier decision stages. This strategy alleviates the credit assignment problem in multi-stage decisions and facilitates stable optimization of policies that jointly improve diagnostic performance and cost efficiency.

\subsection{Problem Formulation}

In this study, we consider multi-omics data classification as a representative medical application scenario of our general staged decision-making method. Suppose there are $M$ heterogeneous omics modalities available in clinical practice. The diagnostic process is modeled as a finite-stage procedure involving up to $M$ modality acquisition decisions, and the maximum decision horizon is therefore set to $T=M$. Given a dataset containing $N$ subjects, each subject is represented by $M$ modality-specific feature vectors and a corresponding class label, denoted as $\{x_i^{(1)},x_i^{(2)},\ldots,x_i^{(M)},y_i\}$, $s.t.\ i\in\{1,\ldots,N\}$, where $x_i^{(m)}\in\mathbb{R}^{d_m}$ represents the feature vector of the $m$-th modality for the $i$-th subject, and $y_i\in\{1,\ldots,C\}$ denotes the ground-truth class label, with $C$ being the number of classes.

The modality-aware classification process is formulated as a finite-horizon MDP, where the decision horizon coincides with the number of available modalities. At each stage $t$, SDM-Q determines whether to acquire an additional modality based on the current diagnostic state. This decision is guided by a learned action--value function $Q(S_t,a_t)$, which balances the expected diagnostic benefit of candidate modalities against their corresponding acquisition costs. Through this formulation, the model learns a subject-specific dynamic modality selection policy.

Formally, the MDP is defined by the five-tuple $(S,A,P,R,\gamma)$. Specifically, $S$ denotes the state space representing different combinations of acquired modalities, $A$ denotes the discrete action space comprising the \textit{Stop} action and the distinct actions of acquiring each of the remaining unacquired candidate modalities, $P$ represents the state transition mechanism determined by modality acquisition, $R$ denotes the reward function defined by classification correctness and modality-specific costs, and $\gamma\in(0,1]$ is the discount return factor. The detailed definitions of these components are described below.

\textbf{State representation.} To accommodate dynamically varying modality combinations while maintaining a fixed input dimensionality, we introduce a modality mask mechanism. Let $M_t \subseteq \{1,2,\ldots,M\}$ denote the set of modalities that have been acquired at stage $t$, where $|M_t|=t$. For modalities that have not yet been acquired, their feature vectors are replaced with zeros. Accordingly, the state representation of the $i$-th subject at stage $t$ is defined in Eq.~\eqref{eq:state}.

\begin{equation}
S_{i,t} =
\left[
x_i^{(1)}\mathbb{I}(1 \in M_{i,t}),
x_i^{(2)}\mathbb{I}(2 \in M_{i,t}),
\ldots,
x_i^{(M)}\mathbb{I}(M \in M_{i,t})
\right],
\label{eq:state}
\end{equation}

\noindent where $I(\cdot)$ denotes the indicator function indicating whether the corresponding omics data has been acquired.

\textbf{Action.} At each decision stage $t$, SDM-Q selects an action $a_t$ from a discrete action space $a_t=\{0\}\cup U_t$, where $U_t=\{1,2,\ldots,M\}\backslash M_t$ represents the set of remaining unacquired candidate modalities. Specifically, $a_t=0$ denotes the \textit{Stop} action, while $a_t=m\in U_t$ corresponds to the distinct action of acquiring a specific modality $m$. If an acquisition action is selected $(a_t=m)$, the modality set is updated as $M_{t+1}=M_t\cup\{m\}$, and the process proceeds to the next decision stage, i.e. $S_{t+1}$. If the \textit{Stop} action is executed $(a_t=0)$, or when the maximum stage $t_{\max}=M$ is reached, the decision process terminates and the final classification prediction is produced.

\textbf{State transition.} At each decision step, SDM-Q updates its state according to the current state $S_t$ and the selected action $a_t$. If an acquisition action is chosen, i.e., $a_t=m\in U_t$, the process advances to the subsequent stage. If the \textit{Stop} action is selected, i.e., $a_t=0$, the state transitions directly to a terminal state, and the classification process ends with the final prediction output.

\textbf{Modality cost.} Each modality $m$ is associated with a non-negative acquisition cost $C_m\geq0$. If the modality acquired at stage $t$ for subject $i$ is denoted as $m_t^{(i)}$, its incremental cost is $\mathrm{Cost}(M_t^{(i)})$. Accordingly, the cumulative acquisition cost incurred up to stage $t$ is defined as in Eq.~\eqref{eq:cost}.

\begin{equation}
\mathrm{Cost}(M_{i,t})
=
\sum_{k=1}^{t} C_{m_{i,k}}.
\label{eq:cost}
\end{equation}

\textbf{Reward design.} In this study, rewards are provided only when the \textit{Stop} action is executed. To encourage accurate classification under cost constraints, we adopt a sparse binary reward mechanism based on classification outcomes. This design focuses the agent on learning reliable decision boundaries while reducing value estimation instability caused by probability calibration errors during early training. Let $\hat{y}_i$ denote the predicted label of subject $i$. The classification correctness is mapped to a signed reward as shown in Eq.~\eqref{eq:rewardsign}.

\begin{equation}
\mathrm{RewardSign}(y_i,\hat{y}_i)=
\begin{cases}
+1, & \hat{y}_i=y_i,\\
-1, & \hat{y}_i\neq y_i,
\end{cases}
\label{eq:rewardsign}
\end{equation}

\noindent which converts classification outcomes into positive or negative reinforcement reward signals. Accordingly, the reward obtained when the \textit{Stop} action is executed at stage $t$ is defined as shown in Eq.~\eqref{eq:stopreward}.

\begin{equation}
R(S_{i,t},\mathrm{Stop})
=
\mathrm{RewardSign}(y_i,\hat{y}_i)
-
Cost(M_{i,t}).
\label{eq:stopreward}
\end{equation}

In contrast, an acquisition action $(a_t=m\in U_t)$ does not yield an immediate reward, as shown in Eq.~\eqref{eq:zeroreward}.

\begin{equation}
R(S_{i,t}, m) = 0,\quad \forall m \in U_t .
\label{eq:zeroreward}
\end{equation}

An acquisition action strictly expands the available modalities without generating a classification output. We assign a zero immediate reward to this action rather than a step penalty to prevent double-counting, as cumulative acquisition costs are already deducted at the terminal stage shown in Eq.~\eqref{eq:stopreward}. Crucially, this delayed reward mechanism synergizes with our backward stage-wise training. Because future state values estimated by downstream networks inherently encode the optimal trade-off between predictive accuracy and cumulative costs, a zero immediate reward compels the agent to optimize the global objective. This precludes short-sighted policies that prematurely terminate the classification process merely to avoid step penalties.

\textbf{Objective function.} Based on the above definitions of states, actions, and rewards, the dynamic multi-omics acquisition process is formulated as a finite-horizon RL problem. The objective is to learn an optimal policy $\pi^{*}$ that maximizes the expected cumulative return, as defined in Eq.~\eqref{eq:objective}.

\begin{equation}
\pi^*
=
\arg\max_{\pi}
\mathbb{E}_{\pi}
\left[
\sum_{t=1}^{T}
\gamma^{t-1}
R(S_{i,t},a_{i,t})
\right],
\label{eq:objective}
\end{equation}

\noindent where $\gamma\in(0,1]$ denotes the discount factor, and $R(S_t^{(i)},a_t^{(i)})$ represents the reward obtained after executing action $a_t$ at state $S_t$ for subject $i$ at stage $t$.

\subsection{Cost-Aware Deep Q-Learning}

To achieve an optimal trade-off between modality acquisition cost and classification accuracy, we adopt the Deep Q-Learning algorithm. Our sequential decision-making framework operates within a strictly discrete action space consisting of \textit{Acquisition} and \textit{Stop} actions, making Q-learning highly suitable for estimating the value distributions of such discrete choices \cite{ref6}. The core objective is to learn an action--value function $Q(S_t,a_t)$, which estimates the maximum expected cumulative discounted return obtained by executing action $a_t$ at state $S_t$. Based on the reward formulation defined in Section~3.1, the target Q-values for training are constructed according to the Bellman equation \cite{ref12}.

\textbf{Action--value function.}
For any state--action pair $(S_t,a_t)$, the action--value function under policy $\pi$ is defined as in Eq.~\eqref{eq:qvalue}.

\begin{equation}
Q_{\pi}(S_{i,t},a_{i,t})
=
\mathbb{E}_{\pi}
\left[
\sum_{\tau=t}^{T_i}
\gamma^{\tau-t}
R(S_{i,\tau},a_{i,\tau})
\right],
\label{eq:qvalue}
\end{equation}

\noindent where $T_i$ denotes the terminal stage of subject $i$. The optimal action--value function satisfies the Bellman optimality equation, as shown in Eq.~\eqref{eq:bellman}.

\begin{equation}
Q_{\pi}^{*}(S_{i,t},a_{i,t})
=
R(S_{i,t},a_{i,t})
+
\gamma
\max_{a'}
Q^{*}(S_{i,t+1},a'),
\label{eq:bellman}
\end{equation}

\textbf{Target for the Stop action.}
When the agent executes the \textit{Stop} action $(a_t=\textit{Stop})$, the decision process terminates and the target Q-value is directly determined by the terminal reward, as denoted in $y_{i,t} = R(S_{i,t},\textit{Stop})$, which is defined in Eq. \eqref{eq:stopreward}. Then, the Q-learning loss associated with the \textit{Stop} action is defined as in Eq. \eqref{eq:stoploss}.

\begin{equation}
L_Q^{(\textit{stop})}
=
\frac{1}{|B|}
\sum_{i=1}^{|B|}
\left(
Q_{\pi}^{*}(S_{i,t},a_{i,t})
-
y_{i,t}^{(\textit{stop})}
\right)^2.
\label{eq:stoploss}
\end{equation}

\noindent where $B$ denotes the minibatch of subjects randomly sampled from the training set, and $|B|$ represents the batch size.

\textbf{Target for Acquisition actions.}
An acquisition action $(a_t=m\in U_t)$ does not yield an immediate reward. Executing this action deterministically transitions the current state $S_t$ to the subsequent state $S_{t+1}$ by incorporating the newly acquired modality $m$. The target Q-value for an acquisition action is computed based on the maximal expected return at the next stage, as defined in Eq.~\eqref{eq:futurevalue}.

\begin{equation}
V_{i,t}^{(m)}
=
\max_{a'\in\{0\}\cup U_{t+1}}
Q_{\theta^{-}}(S_{i,t+1},a'),
\label{eq:futurevalue}
\end{equation}
\noindent where $\theta^{-}$ denotes the parameters of the frozen target network.

The target Q-value for an \textit{Acquisition} action is defined as in Eq.~\eqref{eq:acquisitiontarget}.

\begin{equation}
y_{i,t}^{(\textit{acquisition})}
=
\gamma
\max_{m\in U_t}
V_{i,t}^{(m)}.
\label{eq:acquisitiontarget}
\end{equation}

If $U_t=\varnothing$, we set $y_{t,\textit{acquisition}}=0$ by convention. Then, the corresponding Q-learning loss is defined in Eq.~\eqref{eq:acquisitionloss}.

\begin{equation}
L_Q^{(\textit{acquisition})}
=
\frac{1}{|B|}
\sum_{i=1}^{|B|}
\left(
Q_{\theta}(S_{i,t},m)
-
y_{i,t}^{(\textit{acquisition})}
\right)^2.
\label{eq:acquisitionloss}
\end{equation}

\textbf{Classification loss.}
The classification head is jointly trained on the shared feature representation. Specifically, for a subject $i$ at stage $t$, this representation is denoted as $u_t^{(i)}$, which is extracted from the subject's current state $S_t^{(i)}$ by the modality encoders of the the stage-specific decision network. The classification loss over a mini-batch $B$ is defined using the cross-entropy, as shown in Eq.~\eqref{eq:classificationloss}.

\begin{equation}
L_{clf}
=
-
\frac{1}{|B|}
\sum_{i=1}^{|B|}
\sum_{c=1}^{C}
\mathbb{I}(y_i=c)
\log
p_{\theta}
(c\mid u_{i,t}),
\label{eq:classificationloss}
\end{equation}

\noindent where $p_{\theta}(\cdot)$ denotes the softmax output of the classification head.
\subsection{Stage-wise Decision Network Architecture}

\begin{figure}[!htbp]
\centering
\includegraphics[width=0.95\textwidth]{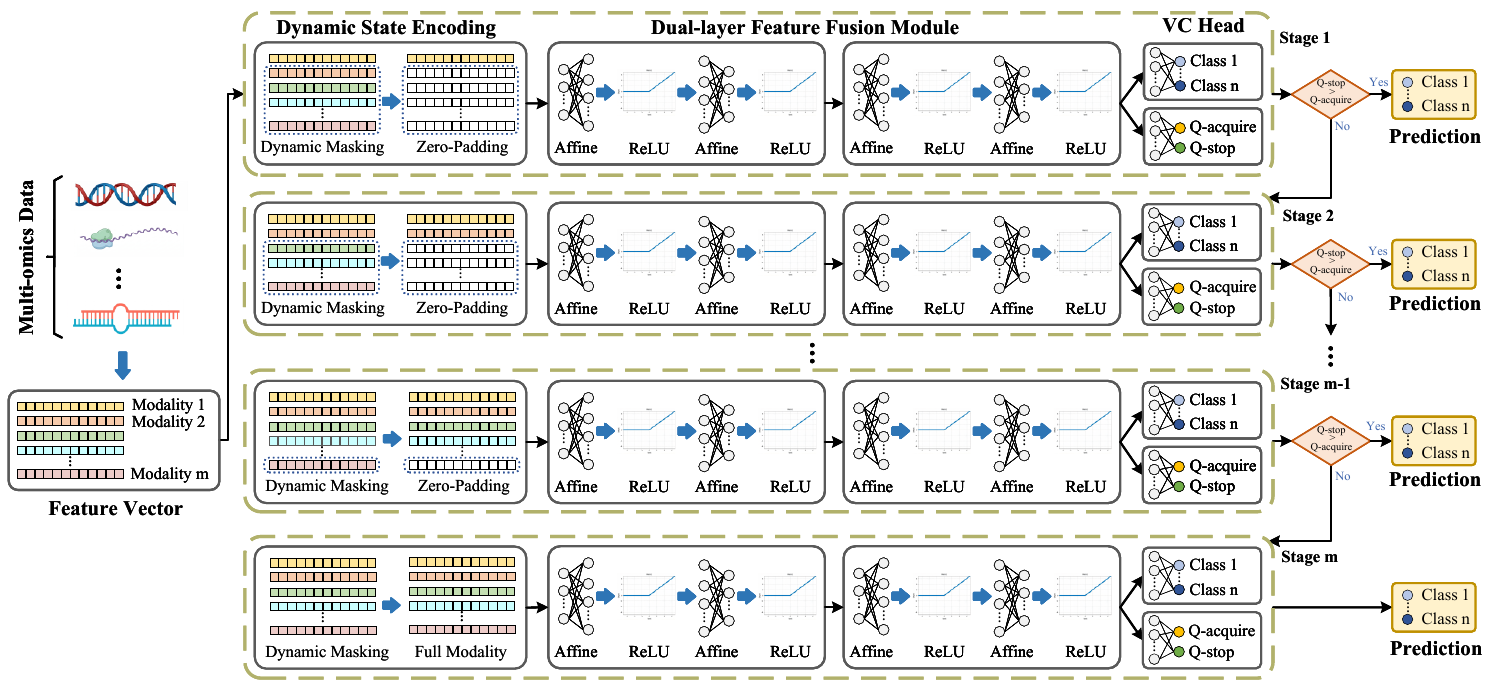}
\caption{
Stage-wise computational framework of SDM-Q.
}
\label{fig:mdpm}
\end{figure}

To support modality-level sequential acquisition and multi-omics classification, we propose SDM-Q, a staged dynamic decision architecture. As illustrated in Fig.~\ref{fig:mdpm}, for each subject, the raw multi-omics data are first transformed into modality-specific feature vectors, which serve as candidate modality representations for subsequent staged decision-making during the training. The feature vector of the $i$-th subject in the $m$-th omics modality is denoted as $\mathbf{x}_i^{(m)}$. At each decision stage, SDM-Q constructs a stage-specific state according to the modalities acquired so far and forwards this state to the corresponding stage-specific decision module. Although the modules at different stages share the same computational architecture, their parameters are learned independently, allowing each module to model the state distribution and decision function under a specific modality-availability condition. Overall, SDM-Q comprises three core components: dynamic state encoding, dual-layer feature fusion, and a value-classification head. By integrating these components, SDM-Q establishes a unified decision-making framework that jointly supports stage-wise state representation, cross-modality information fusion, action-value estimation over the current decision space, and classification prediction.

\textbf{Dynamic State Encoding Module.}
This module constructs stage-specific multi-omics states during sequential modality acquisition through dynamic masking and zero-padding. Since SDM-Q does not assume that all omics modalities are available at the initial stage, the state representation is progressively updated according to the modalities acquired along the decision process. This design preserves a consistent input structure across stages while explicitly representing the modality availability of each sample. For the $m$-th omics modality of the $i$-th sample at decision stage $t$, the masked stage input is defined in Eq. \eqref{eq:mask}.

\begin{equation}
\tilde{\mathbf{x}}_{i,t}^{(m)}
=
I(m\in M_{i,t})\mathbf{x}_{i}^{(m)},
\label{eq:mask}
\end{equation}
where $I(\cdot)$ denotes the indicator function. If $m\in M_{i,t}^{(i)}$, it indicates that the modality $m$ has been acquired and its original feature vector is retained; otherwise, the corresponding input is replaced with a zero vector. The stage-specific state of the $i$-th sample is then formulated as shown in Eq. \eqref{eq:stage_specific_state}.
\begin{equation}
S_{i,t}
=
\left[
\tilde{\mathbf{x}}_{i,t}^{(1)},
\tilde{\mathbf{x}}_{i,t}^{(2)},
\ldots,
\tilde{\mathbf{x}}_{i,t}^{(M)}
\right].
\label{eq:stage_specific_state}
\end{equation}

Through dynamic masking and zero-padding, $S_{i,t}$ maintains a fixed input dimensionality across decision stages while retaining only the effective information from the modalities acquired so far. When the policy selects an acquisition action $\textit{Acquire}(m)$, the corresponding modality is activated in the next stage and incorporated into the updated state representation. When the selected action is \textit{Stop}, the classification prediction produced from the current state is used as the final output. Under the finite-stage setting, if a sample reaches the final stage, all candidate modalities are incorporated into the state representation, and $S_{i,t}$ becomes equivalent to the complete multi-omics input.

\textbf{Dual-layer Feature Fusion Module.}
After the stage-specific state $S_{i,t}$ is constructed, this module maps the modality-level masked inputs into a shared latent representation for subsequent action-value estimation and classification prediction. The transformation is organized in two successive stages. The first stage performs modality-specific encoding, where each masked modality input is projected into a low-dimensional embedding space. The second stage performs cross-modality fusion by transforming the concatenated modality embeddings into a shared representation for the current decision stage.

For the $m$-th omics modality of the $i$-th subject at decision stage $t$, the masked input $\tilde{\mathbf{x}}_{i,t}^{(m)}$ is encoded as
\begin{equation}
\mathbf{h}_{i,t}^{(m)}
=
f_{\mathrm{enc}}^{(m)}
\left(
\tilde{\mathbf{x}}_{i,t}^{(m)}
\right),
\end{equation}
where $f_{\mathrm{enc}}^{(m)}(\cdot)$ denotes the modality-specific encoding mapping for modality $m$, which projects the masked modality input into a low-dimensional embedding space. In our implementation, this mapping is implemented using two affine transformations with ReLU nonlinearities. The resulting vector $\mathbf{h}_{i,t}^{(m)}$ represents the modality-specific embedding at the current decision stage.

The modality embeddings are then arranged according to predefined modality index positions and concatenated to form the multi-omics joint representation:
\begin{equation}
\mathbf{z}_{i,t}
=
\mathrm{Concat}
\left(
\mathbf{h}_{i,t}^{(1)},
\mathbf{h}_{i,t}^{(2)},
\ldots,
\mathbf{h}_{i,t}^{(M)}
\right).
\end{equation}
The predefined modality index positions are used only to maintain a consistent representation layout and do not impose a fixed modality acquisition order across subjects. The actual acquisition path is determined adaptively by the learned stage-wise policy according to the current state and the estimated action values over the valid action set.

The joint representation $\mathbf{z}_{i,t}$ is further transformed by a shared fusion mapping:
\begin{equation}
\mathbf{u}_{i,t}
=
f_{\mathrm{fus}}
\left(
\mathbf{z}_{i,t}
\right),
\end{equation}
where $f_{\mathrm{fus}}(\cdot)$ denotes the shared fusion mapping applied to the concatenated multi-omics representation. This mapping converts the joint modality embedding into a cross-modality shared representation and is also implemented using two affine transformations with ReLU nonlinearities. The output $\mathbf{u}_{i,t}$ serves as the stage-specific fused representation and is fed in parallel into the action-value estimation branch over the current decision space and the classification prediction branch, thereby providing a common feature basis for policy learning and diagnostic prediction.

\textbf{Value--Classification Head.}
After obtaining the shared fused representation $\mathbf{u}_{i,t}$, SDM-Q employs a parallel value--classification head to jointly support action-value estimation and class-probability prediction. The shared representation is not explicitly decomposed into separate task-specific feature subspaces. Instead, it is fed into both the action-value branch and the classification branch, allowing sequential modality acquisition and diagnostic prediction to be learned from a consistent stage-specific state representation.

\textbf{1) Action-value branch (Q-head).}
The action-value branch takes $\mathbf{u}_{i,t}$ as input and outputs an action-value vector over the valid decision space at the current stage. The valid action set is defined as
\begin{equation}
{A}_{i,t}^{\mathrm{valid}}
=
\{\textit{Stop}\}
\cup
\{\textit{Acquire}(m)\mid m\in U_{i,t}\},
\end{equation}
where $U_{i,t}$ denotes the set of modalities that have not yet been acquired for subject $i$ at stage $t$. Accordingly, the Q-head estimates the expected return of terminating the acquisition process as well as the expected returns of acquiring specific candidate modalities:
\begin{equation}
\mathbf{q}_{i,t}
=
Q_{\theta_t}(S_{i,t})
=
\left[
Q_{\theta_t}(S_{i,t},\textit{Stop}),
\left\{
Q_{\theta_t}(S_{i,t},\textit{Acquire}(m))
\right\}_{m\in U_{i,t}}
\right].
\end{equation}
Here, $Q_{\theta_t}(S_{i,t},\textit{Stop})$ represents the value of terminating the decision process and performing classification based on the currently observed modalities, whereas $Q_{\theta_t}(S_{i,t},\textit{Acquire}(m))$ represents the value of acquiring a specific unobserved modality $m$ and proceeding to the next decision stage. This branch is optimized through the Q-learning loss $L_Q$ and provides the value basis for learning the dynamic modality acquisition and stopping policy.

\textbf{2) Classification branch.}
The classification branch maps the shared representation $\mathbf{u}_{i,t}$ to the class-probability distribution $p_{\theta_t}(c\mid S_{i,t})$. This branch is optimized using the cross-entropy loss $L_{\mathrm{clf}}$, enabling the model to produce diagnostic predictions under different partially observed modality states. During inference, when the selected action is \textit{Stop}, the classification output produced from the current state is used as the final diagnostic prediction.

At each decision stage, SDM-Q selects the action with the highest estimated value from the valid action set, denoted as $\text{argmax}_{a_t} \mathbf{q}_{i,t}$. Invalid acquisition actions corresponding to already acquired modalities are masked before action selection. If the selected action is \textit{Stop}, the model terminates further modality acquisition and outputs the predicted class distribution. If the selected action is $\textit{Acquire}(m)$, modality $m$ is incorporated into the observed modality set and the process proceeds to the next stage. Through this mechanism, the value--classification head enables SDM-Q to jointly model subject-specific sequential modality acquisition and diagnostic classification within a unified representation framework.

In summary, SDM-Q integrates dynamic state encoding, modality-level feature transformation, dual-layer feature fusion, and joint value-classification modeling into a unified staged decision framework. By maintaining a consistent state representation and a shared feature learning process under different modality availability conditions, the architecture supports the coordinated optimization of reinforcement learning-based action-value estimation and the classification objective. This design provides a stable and scalable foundation for learning sample-specific dynamic modality acquisition strategies.

\subsection{Backward Stage-wise Training}

To address reward sparsity and credit assignment in multi-step decision-making tasks~\cite{ref13}, we introduce a backward stage-wise training strategy. Unlike conventional approaches that jointly optimize all decision stages, the proposed strategy adopts a recursive optimization paradigm in which value estimates are propagated from the terminal stage to the initial stage. By fixing the value estimates of later-stage modules during training, this strategy alleviates the non-stationary target problem commonly encountered in Q-learning and improves the stability of policy learning. The decision policy is implemented by a sequence of stage-specific decision modules, which contains the weights at different stages. For clarity, the module operating at stage $t$ is denoted as ${D}_t$, and the overall optimization procedure is summarized in Algorithm~\ref{alg:training}.

During training, stage-specific states are constructed for each decision module. For the module at stage $t$, a modality mask is randomly generated such that the number of activated modalities satisfies $|M_{i,t}|=t$. This design simulates diverse modality-availability conditions during training, enabling the model to learn decision patterns under different modality combinations.

Training begins from the terminal stage, denoted as $t=N_{\mathrm{stage}}$, where all modalities have already been acquired and the decision space degenerates into the \textit{Stop} action. At this stage, the action value is directly determined by the terminal reward. The module first computes the classification logits through the classification branch, and the predicted label is obtained by the $\operatorname{argmax}$ operation. The reward is then calculated by combining classification correctness with the accumulated modality acquisition cost. Specifically, the terminal reward $R(S_{i,t}, \textit{Stop})$ is defined as a signed indicator of prediction correctness minus the total cost of the acquired modalities. This terminal stage provides a stable value anchor for backward value propagation, thereby stabilizing the learning of earlier decision stages.

During backward training, once the subsequent module ${D}_{t+1}$ has been trained, its parameters are frozen and training proceeds to the preceding module ${D}_{t}$. At non-terminal stages, the module estimates the values of both the \textit{Stop} and $\textit{Acquire}(m)$ decisions. For the \textit{Stop} decision, the value target follows the same formulation as the terminal stage and depends on the diagnostic reward under the current state. For $\textit{Acquire}(m)$ decision, the long-term return is estimated by considering the future value after acquiring additional omics information and transitioning to the next decision stage.

During backward training, once the subsequent module ${D}_{t+1}$ has been trained, its parameters are frozen and training proceeds to the preceding module ${D}_{t}$. At non-terminal stages, the module estimates the values of both the \textit{Stop} and \textit{Acquire}(m) decisions. For the \textit{Stop} decision, the value target follows the same formulation as the terminal stage and depends on the diagnostic reward under the current state. For the \textit{Acquire}(m) decision, the target value is constructed by considering the future value of the next-stage states obtained after acquiring one additional modality.

Specifically, for each currently unavailable modality $m\in U_{i,t}$, activating this modality generates a candidate next-stage state $S_{i,t+1}^{(m)}$. The frozen downstream module ${D}_{t+1}$ is then used to estimate the maximum future return under this candidate state. The target for the current \textit{Acquire}(m) decision is defined by selecting the candidate modality that yields the highest downstream value and applying the discount factor. This formulation enables the current-stage module to learn whether continuing the acquisition process is beneficial, while leaving the specific next-modality selection to the downstream value evaluation.

This backward value propagation strategy allows earlier decision modules to be optimized using stable value estimates derived from later stages, thereby improving policy learning stability and facilitating an effective trade-off between diagnostic accuracy and modality acquisition cost.
The target Q-value is defined according to the current decision and whether the process has reached the terminal stage:
\begin{equation}
y_{i,t}(a_t)=
\begin{cases}
R(S_{i,t},a_t),
& a_t=\textit{Stop}\ \text{or}\ t=M,
\\[4pt]
R(S_{i,t},a_t)
+\gamma
\max\limits_{m\in {U}_{i,t}}
\max\limits_{a'\in\{0,1\}}
Q_{\theta^-}\left(S_{i,t+1}^{(m)},a'\right),
& a_t=\textit{Acquire}(m),\ t<M.
\end{cases}
\label{eq:targetq}
\end{equation}

where $S_{i,t+1}^{(m)}$ denotes the candidate next state obtained by activating modality $m$. Since the \textit{Acquire}(m) action does not directly produce a classification output, its immediate reward is set to zero in this study, i.e., $R(S_{i,t},\textit{Acquire}(m))=0$. Therefore, the target value of the \textit{Acquire}(m) action is mainly determined by the discounted optimal future value estimated by the frozen downstream module.

The final loss function jointly optimizes the Q-learning objective and the classification objective, as shown in Eq. \eqref{eq:finalloss}.

\begin{equation}
L=L_Q+\lambda_{clf}L_{clf}.
\label{eq:finalloss}
\end{equation}

\noindent where $L_Q$ is used to train the action values of \textit{Acquire}(m) and \textit{Stop}, and $L_{clf}$ is used to optimize the classification branch so that the model can generate reliable predictions when the \textit{Stop} action is selected. Through this joint optimization scheme, the model simultaneously minimizes Q-value estimation error and classification error, enabling a balanced optimization of modality acquisition cost and classification accuracy. The detailed backward stage-wise optimization procedure for training the stage-specific decision modules is summarized in Algorithm~\ref{alg:training}.

\begin{algorithm}[H]
\small
\caption{Backward Stage-wise Training for Stage-specific Decision Networks}
\label{alg:training}

\Input{
Modality features $X=\{x^{(1)},x^{(2)},\ldots,x^{(M)}\}$,
labels $y$, modality cost map $C_m$, training epochs $E$
}

\Output{
Trained networks $D_{\theta 1},\ldots,D_{\theta M}$
}

\For{$m=1$ to $M$}{
    Randomly initialize $D_{\theta m}$\;
}

\For{$t=M$ down to $1$}{
    Freeze parameters of $D_j$ for all $j>t$\;

    \For{\text{epoch} $e=1$ to $E$}{
        Sample a minibatch $B$\;

        \For{each subject $i$ in $B$}{
            Randomly generate modality mask $\mathrm{mask}_i$\;

            Construct state:
            $S_{i,t}
            =
            [x_i^{(1)}\mathbb{I}(1\in M_{i,t}),
            x_i^{(2)}\mathbb{I}(2\in M_{i,t}),
            \ldots,
            x_i^{(M)}\mathbb{I}(M\in M_{i,t})]$\;

            Forward $D_{\theta t}$ to obtain
            $Q_{\theta}(S_t^{(i)},a^{(i)})$
            and predicted probability $\hat{p}_i$\;

            Compute reward $R_i$\;

            \eIf{$t=M$}{
                $y_i(\textit{stop})=R_i$\;
            }{
                \For{\text{each unacquired modality} $m\in U_{t}$}{
                    Construct next state $S_{t+1}^{(i)}$\;

                    $V_{i,t}^{(m)}
                    =
                    \max\limits_{a'}
                    Q_{\theta^-}
                    \left(
                    S_{i,t+1}^{(m)},a'
                    \right)$\;
                }

                $y_i(\textit{Acquire}(m))=\gamma V_{i,t}^{(m)}$\;
            }
        }

        Compute losses $L_Q$ and $L_{\mathrm{clf}}$\;

        Update parameters of $D_{\theta t}$\;
    }
}

\end{algorithm}

Overall, the proposed backward stage-wise training strategy allows earlier decision stages to leverage reliable future value estimates obtained from later stages. By providing fixed learning targets, this mechanism significantly improves the stability of dynamic modality selection policies.

\subsection{Inference and Staged Decision Process}

The objective of the inference stage is to adaptively determine, for each subject, both the number of required modalities and their acquisition order based on the trained stage-specific value functions. Unlike the training stage, which employs randomly generated modality masks, the inference stage adopts a deterministic greedy value-based decision strategy and allows any single modality to serve as the initial state. The overall inference procedure is summarized in Algorithm~2.

Specifically, an initial modality is first assigned to each subject, with its index denoted as 
\(m_{i,0} \in \{1,2,\ldots,M\}\). The corresponding initial modality mask is defined as:

\begin{equation}
\mathrm{mask}_{i,1}(m)=
\begin{cases}
1, & m=m_{i,0}, \\
0, & m\neq m_{i,0}.
\end{cases}
\label{eq19}
\end{equation}
Accordingly, the initial state contains only the features of modality $m_0$, while all other modalities remain locked. At stage $t$, the set of acquired modalities is defined as $M_t=\{m \mid \mathrm{mask}_t(m)=1\}$. The corresponding stage-specific network $\mathrm{D}_t$ outputs the action values corresponding to the available action space $A_t=\{0\}\cup U_t$. The inference process terminates when either of the following conditions is satisfied: (i) the \textit{Stop} action yields the highest Q-value, i.e., or (ii) the maximum stage is reached. If the process does not terminate, the agent greedily selects the acquisition action to acquire the next optimal modality and transitions to stage $t+1$.

At termination, the final predicted class label is produced by the classification head:

\begin{equation}
\hat{y}=\arg\max_c\ p_\theta(y=c\mid S_t).
\label{eq20}
\end{equation}

If the inference process does not terminate, the model selects the next optimal modality $m^{*}$ directly based on the current state $S_t$. Since the Q-network is optimized to estimate the long-term return, it greedily selects the candidate modality with the maximal Q-value from the unacquired set $U_t$:

\begin{equation}
m^{*}=\arg\max_{m\in U_t}Q_\theta(S_t^{(m)},m).
\label{eq21}
\end{equation}

The modality mask and state representation are subsequently updated, and the decision process proceeds to the next stage. The greedy value-based inference procedure for adaptive modality selection is summarized in Algorithm~\ref{alg:inference}.

\FloatBarrier

\begin{algorithm}[!htbp]
\caption{Greedy Value-based Inference for Adaptive Modality Selection}
\label{alg:inference}

\Input{
Modality features $X=\{x^{(1)},x^{(2)},\ldots,x^{(M)}\}$,
trained networks $D_{\theta 1},\ldots,D_{\theta M}$,
initial modality $m_0$
}

\Output{
Predicted label $\hat{y}$
}

Initialize $\mathrm{mask}_1$ with $\mathrm{mask}_1(m_0)=1$, others $0$\;

Construct initial state $S_0$\;

$t=0$\;

\While{True}{
    $k=|M_t|$\;

    Compute Q-values $Q_{\theta}(S_t^{(i)},a)$ for all valid action
    $a\in\{0\}\cup U_t$\;

    \If{
    $Q_{\theta}(S_t^{(i)},0)\geq Q_{\theta}(S_t^{(i)},m)$
    \textbf{or}
    $|M_t|=3$
    }{
        $\hat{y}=\arg\max_c p_{\theta}(c\mid S_t^{(i)})$\;

        $K=|M_t|$\;

        Break\;
    }

    Greedily select next modality:
    $m^{*}=\arg\max_{m\in U_t}Q_{\theta}(S_t^{(i)},m)$\;

    Update $\mathrm{mask}_{t+1}^{(i)}(m^{*})=1$\;

    Construct state $S_{t+1}$\;

    $t=t+1$\;
}

Return $\hat{y},K$\;

\end{algorithm}

Overall, this inference procedure employs a deterministic value-based greedy strategy to enable subject-specific adaptive selection of both modality order and modality quantity. Moreover, it remains fully consistent with the value function learned during training, allowing the trade-off between acquisition cost and diagnostic benefit to be naturally reflected during inference.
\section{Experiments}
\label{sec:experiments}

\subsection{Datasets and Modalities}

\textbf{Datasets.}
We evaluate the proposed method on four publicly available multi-omics benchmark datasets: ROSMAP, LGG, BRCA, and KIPAN. These datasets cover diagnostic tasks for neurodegenerative diseases and molecular subtyping of multiple cancers, exhibiting substantial diversity in terms of subject size, class configuration, and biological heterogeneity. Consequently, they are widely adopted for evaluating multi-omics learning approaches.

ROSMAP~\cite{ref14}: The ROSMAP dataset is designed for Alzheimer's disease diagnosis and contains 351 subjects, including both patients and healthy controls. The task is formulated as a binary classification problem. Due to the relatively limited subject size and heterogeneous biological signals across modalities, ROSMAP is commonly used to evaluate the robustness of multi-omics models under complex biological conditions.

LGG~\cite{ref15}: The LGG dataset focuses on glioma grade classification and contains 510 subjects. Compared with ROSMAP, LGG involves more subtle biological differences between classes, which often requires the integration of multiple omics modalities to achieve reliable classification performance.

BRCA~\cite{ref16}: The BRCA dataset is used for breast cancer molecular subtype classification and includes 876 subjects in a multi-class setting. Owing to its relatively large scale, this dataset is widely used to evaluate the performance of multi-omics models in large-scale multi-class scenarios.

KIPAN~\cite{ref16}: The KIPAN dataset contains 658 subjects from multiple renal cancer subtypes. In this dataset, certain omics modalities exhibit relatively strong discriminative signals, making it suitable for analyzing the relative contributions of different modalities in cancer subtype classification.

\textbf{Modality Description.}
Across all datasets, each subject contains three heterogeneous omics modalities: mRNA expression, DNA methylation, and miRNA expression. Specifically, mRNA expression data represent gene-level transcriptional activity, DNA methylation data describe methylation levels at CpG sites, and miRNA expression data capture post-transcriptional regulatory signals through miRNA abundance profiles~\cite{ref14,ref15,ref16}. These three modalities provide complementary biological information from transcriptional activity, epigenetic regulation, and post-transcriptional regulation, respectively. However, these modalities may contain partially redundant information due to underlying biological correlations across different regulatory layers and a staged method is required to lower the costs. In our experiments, the three omics modalities are treated as parallel inputs to construct the multimodal representation used for downstream classification.

\subsection{Experimental Settings and Evaluation Protocol}

A unified experimental pipeline is adopted across all datasets to ensure fair comparison and reproducibility. For each dataset, the proposed model is trained on the training split and evaluated on the corresponding held-out test set provided in MOGONet \cite{ref8}. To reduce the influence of random parameter initialization and stochastic optimization, all experiments are repeated 10 times with different random seeds. The final results are reported as the mean and standard deviation across multiple runs.

\textbf{Experimental Settings.} All models are implemented using the PyTorch (version 2.4.1). Optimization is performed using the Adam optimizer with a learning rate of $1\times10^{-3}$. Unless otherwise specified, all hyperparameters remain consistent across datasets to avoid dataset-specific tuning. The experiments are conducted on a workstation equipped with an NVIDIA GeForce RTX 5070 GPU (12 GB VRAM). To maintain a unified training strategy across datasets with different subject scales, the number of training epochs is determined based on dataset size to ensure stable convergence.

\textbf{Evaluation Protocol.} We evaluate model performance on both binary classification tasks (ROSMAP and LGG) and multi-class classification tasks (BRCA and KIPAN). For binary classification tasks, we report Accuracy, F1-score, and Area Under the Receiver Operating Characteristic Curve (AUC). For multi-class classification tasks, we report Accuracy, F1-score, and Macro-F1.

Beyond standard classification metrics, we further analyze the sequential decision behavior of the proposed method. Specifically, we report the average number of acquired modalities and the distribution of subjects terminating at each decision stage. These statistics provide insights into the model's cost-aware modality acquisition strategy, but they are not used as direct performance metrics for cross-method comparison. All comparisons with baseline methods are conducted based on the final classification predictions obtained in the terminal stage on test set.

To further investigate the influence of modality acquisition cost on decision policies, we predefine several modality cost configurations. For each configuration, the model is trained and evaluated independently. Representative settings are selected to analyze the trade-off between classification performance and acquisition cost in subsequent experiments.

\subsection{Decision Behavior and Cost--Performance Stability Analysis}

\begin{figure}[H]
\centering
\includegraphics[width=\textwidth]{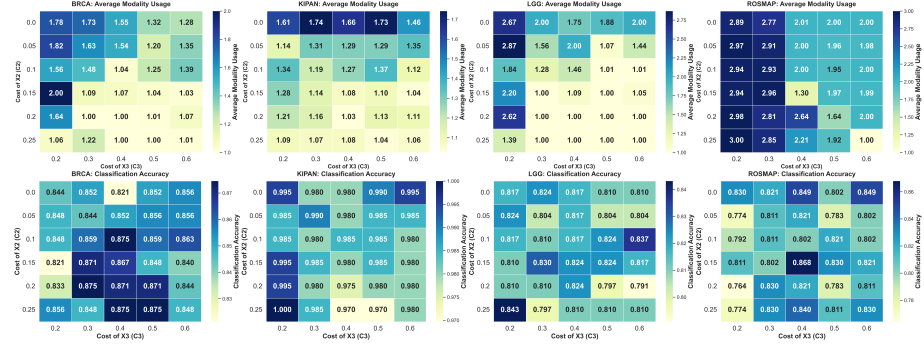}
\caption{Cost sensitivity analysis of SDM-Q under different modality acquisition cost configurations. The top row reports the average number of modalities acquired by the model, while the bottom row shows the  classification accuracy under the corresponding cost settings. From left to right, the columns correspond to BRCA, KIPAN, LGG, and ROSMAP, respectively. Each heatmap cell represents the result obtained under a specific combination of acquisition costs. The horizontal axis denotes the acquisition cost of the third modality, and the vertical axis denotes the acquisition cost of the second modality.}
\label{fig:cost_heatmap}
\end{figure}

We investigate how the proposed framework dynamically balances classification performance with modality acquisition costs under different cost configurations. Specifically, we denote $C_2$ and $C_3$ as the acquisition costs of the second and third modalities, respectively. The first modality (mRNA expression) is treated as the initial input $m_0$ with a fixed acquisition cost of zero. Fig.~\ref{fig:cost_heatmap} presents the cost sensitivity results of SDM-Q on the BRCA, KIPAN, LGG, and ROSMAP datasets, including both the average modality usage and the corresponding classification accuracy under different cost configurations.

\textbf{Adaptive Acquisition Patterns.} The top row of Fig.~\ref{fig:cost_heatmap} illustrates the average modality usage of SDM-Q under varying cost configurations. Across the four datasets, a general decreasing trend in modality usage can be observed as acquisition costs increase. For example, in the BRCA dataset, increasing the cost of the third modality from 0.2 to 0.6 reduces the average modality usage from approximately 1.64 to 1.07. A similar trend is observed in the KIPAN dataset, where the average modality usage decreases from 1.61 to 1.46. These results indicate that under higher acquisition cost constraints, the marginal benefit of acquiring additional modalities becomes less significant. Consequently, the model tends to select the \textit{Stop} action more frequently, leading to earlier termination of the diagnostic process.

ROSMAP shows a stronger reliance on multi-modality integration. Under several cost configurations, its average modality usage remains close to 2 or even higher, suggesting that single-modality information is often insufficient for reliable classification in this task. LGG shows a more transitional response pattern, where the model tends to use more modalities under low-cost settings but substantially reduces modality usage as acquisition costs increase. This finding suggests that the learned acquisition policy is task-dependent, with the balance between cost reduction and classification reliability being adjusted according to the discriminative information available in each dataset.

\textbf{Task-Specific Response Heterogeneity.}
Although increasing acquisition costs consistently suppresses modality usage, the response patterns vary across datasets due to differences in biological characteristics and task difficulty. In the BRCA and KIPAN datasets, the model exhibits strong sensitivity to acquisition costs and quickly converges to single-modality decisions even under relatively low cost levels. This behavior suggests that mRNA expression, acting as the primary modality $m_0$, already provides sufficient discriminative information for reliable classification. In cancer studies such as BRCA and KIPAN, mRNA expression is known to capture essential transcriptomic signatures that are highly indicative of tumor subtypes and biological states, serving as the most informative layer for clinical characterization~\cite{ref17}.

\textbf{Performance Stability under Cost Variations.}
The accuracy heatmaps further show that the classification performance of SDM-Q remains relatively stable under different cost configurations, although the degree of stability varies across datasets. For BRCA, the accuracy ranges from 0.821 to 0.875, indicating that the model maintains comparable performance despite changes in modality acquisition costs. KIPAN shows an even narrower performance range, with accuracy values between 0.970 and 1.000 across all tested configurations. LGG also exhibits moderate stability, with accuracy varying from 0.791 to 0.843. In contrast, ROSMAP presents a relatively larger fluctuation, ranging from 0.764 to 0.868, suggesting that this task is more sensitive to cost settings. Overall, these results indicate that SDM-Q can adjust its modality acquisition behavior under different cost constraints while maintaining broadly comparable classification performance in most cases.

\begin{table}[H]
\centering
\caption{Heterogeneity of diagnostic depth: subject termination proportions and average modality usage across stages.}
\label{tab:termination}

{\fontsize{10pt}{11pt}\selectfont
\setlength{\tabcolsep}{6pt}
\renewcommand{\arraystretch}{0.75}

\begin{tabular*}{\textwidth}{@{\extracolsep{\fill}}cccc@{}}
\toprule
Dataset & Stage & Subjects (\%) & Average modality usage \\
\midrule

\multirow{3}{*}{ROSMAP~\cite{ref14}}
& Stage1 & 2.83\% & \multirow{3}{*}{1.97} \\
& Stage2 & 97.17\% & \\
& Stage3 & 0\% & \\
\midrule

\multirow{3}{*}{LGG~\cite{ref15}}
& Stage1 & 80.39\% & \multirow{3}{*}{1.39} \\
& Stage2 & 0\% & \\
& Stage3 & 19.61\% & \\
\midrule

\multirow{3}{*}{BRCA~\cite{ref16}}
& Stage1 & 99.62\% & \multirow{3}{*}{1.00} \\
& Stage2 & 0.38\% & \\
& Stage3 & 0\% & \\
\midrule

\multirow{3}{*}{KIPAN~\cite{ref16}}
& Stage1 & 95.45\% & \multirow{3}{*}{1.09} \\
& Stage2 & 0.00\% & \\
& Stage3 & 4.55\% & \\
\bottomrule
\end{tabular*}
}

\end{table}
\textbf{Analysis of Classification Depth.} Table \ref{tab:termination} further reports the distribution of subject termination stages under representative cost configurations. In the BRCA and KIPAN datasets, the model achieves high classification efficiency during the initial stage as 99.62\% and 95.45\% of subjects terminate at Stage 1 respectively. This demonstrates that mRNA expression acting as the primary modality $m_0$ provides sufficient information for reliable classification in most breast and renal cancer subjects. In contrast, 97.17\% of ROSMAP subjects proceed to Stage 2, suggesting that additional omics information is often required for accurate diagnosis in this task. Similarly, In the LGG dataset, 80.39\% of subjects terminate at Stage 1, while the remaining subjects require further modality acquisition, reflecting the specific complexity of glioma classification.

Overall, these results demonstrate that SDM-Q can adaptively adjust its modality acquisition strategy according to dataset characteristics and cost constraints, while maintaining broadly comparable classification performance across different cost configurations. This supports its ability to provide cost-aware and subject-specific diagnostic decisions.
\subsection{Performance Comparison with Baseline Methods}

To comprehensively evaluate the effectiveness of the proposed method, we compare SDM-Q with 15 representative classification methods.

We first include several widely used classical baseline methods, including k-nearest neighbors (KNN) ~\cite{ref18}, support vector machines (SVM) ~\cite{ref19}, $\ell_1$-regularized linear regression (LR/Lasso) ~\cite{ref20}, and random forests (RF) ~\cite{ref21}. These methods respectively represent distance-based classification, maximum-margin learning, sparse linear modeling, and ensemble tree-based learning, and have been widely adopted as standard baselines in multi-omics classification studies. In addition, a fully connected neural network (NN) ~\cite{ref22} is included to represent early neural-network-based end-to-end classification models.

To account for statistical learning approaches designed for high-dimensional omics data, we include group-regularized ridge regression (GRridge) ~\cite{ref23}, which introduces group-specific regularization weights based on prior covariate information to improve predictive performance and stability in grouped feature spaces.

We further include block-based multi-omics integration methods, such as block partial least squares discriminant analysis (BPLSDA) ~\cite{ref24} and its sparse extension BSPLSDA (DIABLO framework) ~\cite{ref24}. These methods treat different omics modalities as separate feature blocks and perform integrative analysis by maximizing cross-omics correlations while maintaining discriminative capability.

Among deep multi-omics learning models, we compare with MOGONET ~\cite{ref8}, which constructs modality-specific similarity graphs and employs graph convolutional networks to learn structured representations for each omics modality. The view correlation discovery network (VCDN) module is then used to capture higher-order cross-modality interactions.

To evaluate performance against trustworthy and uncertainty-aware multimodal learning approaches, we include trustworthy multi-view classification (TMC) ~\cite{ref25}, which models evidence strength and uncertainty for each modality using subjective logic and performs evidential fusion across modalities.

We also include several multimodal fusion architectures, including concatenation-based feature fusion (CF) ~\cite{ref26}, gated multimodal units (GMU) ~\cite{ref27}, and multimodal dynamic fusion (MMDynamic) ~\cite{ref28}. These methods adopt different strategies to integrate information across modalities, including late feature fusion, learnable gating mechanisms, and dynamic modality weighting.

Finally, we include discriminative-aware channel pruning (DCP) ~\cite{ref29}, which evaluates the discriminative contribution of feature channels to select informative representations, and cross-omics contrastive learning with self-attention (CLCLSA) ~\cite{ref30}, which aligns multi-omics representations using contrastive learning and models cross-modality dependencies through self-attention mechanisms.

Tables \ref{tab:multiclass} and \ref{tab:binary} summarize the performance of SDM-Q and 15 baseline methods on multi-class classification tasks (BRCA and KIPAN) and binary classification tasks (ROSMAP and LGG).
\begin{table}[H]
\centering
\caption{Performance comparison on multi-class classification tasks (BRCA and KIPAN). The best results are highlighted in bold.}
\label{tab:multiclass}
\resizebox{\textwidth}{!}{%
\begin{tabular}{lcccccc}
\toprule
\multirow{2}{*}{Model} 
& \multicolumn{3}{c}{BRCA~\cite{ref16}} 
& \multicolumn{3}{c}{KIPAN~\cite{ref16}} \\
\cmidrule(lr){2-4} \cmidrule(lr){5-7}
& Accuracy & F1-score & Macro-F1 
& Accuracy & F1-score & Macro-F1 \\
\midrule

KNN~\cite{ref18}      & 0.742$\pm$0.024 & 0.682$\pm$0.025 & 0.730$\pm$0.025 & 0.967$\pm$0.011 & 0.960$\pm$0.014 & 0.967$\pm$0.011 \\
SVM~\cite{ref19}      & 0.729$\pm$0.018 & 0.640$\pm$0.017 & 0.702$\pm$0.017 & 0.995$\pm$0.003 & 0.994$\pm$0.004 & 0.995$\pm$0.003 \\
LR~\cite{ref20}       & 0.732$\pm$0.012 & 0.642$\pm$0.026 & 0.698$\pm$0.026 & 0.974$\pm$0.002 & 0.972$\pm$0.004 & 0.974$\pm$0.002 \\
RF~\cite{ref21}       & 0.754$\pm$0.009 & 0.649$\pm$0.013 & 0.733$\pm$0.013 & 0.981$\pm$0.006 & 0.975$\pm$0.011 & 0.981$\pm$0.006 \\
NN~\cite{ref22}       & 0.754$\pm$0.028 & 0.668$\pm$0.047 & 0.740$\pm$0.047 & 0.991$\pm$0.005 & 0.991$\pm$0.005 & 0.991$\pm$0.005 \\
GRridge~\cite{ref23}  & 0.740$\pm$0.016 & 0.656$\pm$0.025 & 0.726$\pm$0.025 & 0.994$\pm$0.004 & 0.993$\pm$0.004 & 0.994$\pm$0.004 \\
BPLSDA~\cite{ref24}   & 0.642$\pm$0.009 & 0.369$\pm$0.017 & 0.534$\pm$0.017 & 0.933$\pm$0.013 & 0.919$\pm$0.021 & 0.933$\pm$0.013 \\
BSPLSDA~\cite{ref24}  & 0.639$\pm$0.008 & 0.351$\pm$0.022 & 0.522$\pm$0.022 & 0.919$\pm$0.012 & 0.895$\pm$0.014 & 0.918$\pm$0.013 \\
MOGONET~\cite{ref8}   & 0.802$\pm$0.007 & 0.783$\pm$0.010 & 0.718$\pm$0.016 & 0.987$\pm$0.009 & 0.987$\pm$0.009 & 0.986$\pm$0.015 \\
TMC~\cite{ref25}      & 0.842$\pm$0.005 & 0.806$\pm$0.009 & \textbf{0.844$\pm$0.009} & 0.997$\pm$0.003 & 0.994$\pm$0.005 & 0.997$\pm$0.003 \\
CF~\cite{ref26}       & 0.815$\pm$0.008 & 0.771$\pm$0.009 & 0.815$\pm$0.009 & 0.992$\pm$0.005 & 0.988$\pm$0.009 & 0.992$\pm$0.005 \\
GMU~\cite{ref27}      & 0.800$\pm$0.039 & 0.746$\pm$0.058 & 0.798$\pm$0.058 & 0.977$\pm$0.016 & 0.958$\pm$0.032 & 0.976$\pm$0.017 \\
MMDynamic~\cite{ref28}& \textbf{0.861$\pm$0.011} & 0.845$\pm$0.005 & 0.830$\pm$0.015 & 0.999$\pm$0.002 & 0.999$\pm$0.002 & 0.999$\pm$0.003 \\
DCP~\cite{ref29}      & 0.738$\pm$0.043 & 0.743$\pm$0.045 & 0.683$\pm$0.043 & 0.956$\pm$0.017 & 0.957$\pm$0.017 & 0.939$\pm$0.014 \\
CLCLSA~\cite{ref30}   & 0.825$\pm$0.013 & 0.856$\pm$0.006 & 0.831$\pm$0.008 & \textbf{0.999$\pm$0.002} & \textbf{0.999$\pm$0.002} & 0.999$\pm$0.002 \\

\textbf{Proposed} 
              & \textbf{0.861$\pm$0.011} 
              & \textbf{0.863$\pm$0.013} 
              & 0.828$\pm$0.016 
              & 0.998$\pm$0.003 
              & 0.998$\pm$0.003 
              & \textbf{0.999$\pm$0.002} \\
\bottomrule
\end{tabular}
}
\end{table}

\begin{table}[H]
\centering
\caption{Performance comparison on binary classification tasks (ROSMAP and LGG). The best results are highlighted in bold.}
\label{tab:binary}
\resizebox{\textwidth}{!}{%
\begin{tabular}{lcccccc}
\toprule
\multirow{2}{*}{Model} 
& \multicolumn{3}{c}{ROSMAP~\cite{ref14}} 
& \multicolumn{3}{c}{LGG~\cite{ref15}} \\
\cmidrule(lr){2-4} \cmidrule(lr){5-7}
& Accuracy & F1-score & AUC 
& Accuracy & F1-score & AUC \\
\midrule

KNN~\cite{ref18}      & 0.657$\pm$0.036 & 0.671$\pm$0.045 & 0.709$\pm$0.045 & 0.729$\pm$0.034 & 0.738$\pm$0.038 & 0.799$\pm$0.038 \\
SVM~\cite{ref19}      & 0.770$\pm$0.024 & 0.778$\pm$0.026 & 0.770$\pm$0.026 & 0.754$\pm$0.046 & 0.757$\pm$0.046 & 0.754$\pm$0.046 \\
LR~\cite{ref20}       & 0.694$\pm$0.037 & 0.730$\pm$0.035 & 0.770$\pm$0.035 & 0.761$\pm$0.018 & 0.767$\pm$0.027 & 0.823$\pm$0.027 \\
RF~\cite{ref21}       & 0.726$\pm$0.029 & 0.734$\pm$0.019 & 0.811$\pm$0.019 & 0.748$\pm$0.012 & 0.742$\pm$0.010 & 0.823$\pm$0.010 \\
NN~\cite{ref22}       & 0.755$\pm$0.021 & 0.764$\pm$0.025 & 0.827$\pm$0.025 & 0.737$\pm$0.023 & 0.748$\pm$0.037 & 0.810$\pm$0.037 \\
GRridge~\cite{ref23}  & 0.760$\pm$0.034 & 0.769$\pm$0.023 & 0.841$\pm$0.023 & 0.746$\pm$0.038 & 0.756$\pm$0.044 & 0.826$\pm$0.044 \\
BPLSDA~\cite{ref24}   & 0.742$\pm$0.024 & 0.755$\pm$0.025 & 0.830$\pm$0.025 & 0.759$\pm$0.025 & 0.738$\pm$0.023 & 0.825$\pm$0.023 \\
BSPLSDA~\cite{ref24}  & 0.753$\pm$0.033 & 0.764$\pm$0.021 & 0.838$\pm$0.021 & 0.685$\pm$0.027 & 0.662$\pm$0.026 & 0.730$\pm$0.026 \\
MOGONET~\cite{ref8}   & 0.802$\pm$0.015 & 0.813$\pm$0.012 & 0.892$\pm$0.009 & 0.803$\pm$0.024 & 0.807$\pm$0.018 & 0.877$\pm$0.010 \\
TMC~\cite{ref25}      & 0.825$\pm$0.009 & 0.823$\pm$0.006 & 0.885$\pm$0.006 & 0.819$\pm$0.008 & 0.815$\pm$0.004 & 0.871$\pm$0.004 \\
CF~\cite{ref26}       & 0.784$\pm$0.011 & 0.788$\pm$0.005 & 0.880$\pm$0.005 & 0.811$\pm$0.012 & 0.822$\pm$0.004 & 0.881$\pm$0.004 \\
GMU~\cite{ref27}      & 0.776$\pm$0.025 & 0.784$\pm$0.016 & 0.869$\pm$0.016 & 0.803$\pm$0.015 & 0.808$\pm$0.012 & 0.886$\pm$0.012 \\
MMDynamic~\cite{ref28}& 0.806$\pm$0.020 & 0.813$\pm$0.018 & \textbf{0.894$\pm$0.008} & 0.812$\pm$0.010 & \textbf{0.833$\pm$0.011} & 0.868$\pm$0.002 \\
DCP~\cite{ref29}      & 0.703$\pm$0.054 & 0.681$\pm$0.096 & 0.702$\pm$0.057 & 0.730$\pm$0.041 & 0.737$\pm$0.045 & 0.728$\pm$0.040 \\
CLCLSA~\cite{ref30}   & 0.825$\pm$0.013 & 0.824$\pm$0.013 & 0.880$\pm$0.006 & \textbf{0.827$\pm$0.013} & 0.826$\pm$0.011 & 0.894$\pm$0.002 \\

\textbf{Proposed} 
              & \textbf{0.834$\pm$0.025} 
              & \textbf{0.834$\pm$0.025} 
              & 0.893$\pm$0.008 
              & 0.822$\pm$0.011 
              & 0.822$\pm$0.011 
              & \textbf{0.895$\pm$0.007} \\
\bottomrule
\end{tabular}
}
\end{table}

\textbf{Performance of Classical and Statistical Models.}
Conventional machine learning and statistical approaches show clear limitations when applied to high-dimensional multi-omics data. Methods such as KNN, SVM, and RF exhibit limited robustness under noisy and heterogeneous feature distributions due to their shallow modeling capacity. For example, KNN achieves an accuracy of 0.657 and an AUC of 0.709 on the ROSMAP dataset, indicating that distance-based metrics struggle to construct reliable neighborhood structures in high-dimensional omics feature spaces. Statistical learning models incorporating structured priors provide moderate improvements in certain scenarios but remain constrained by linear assumptions. For instance, GRridge achieves an AUC of 0.841 on the ROSMAP dataset, suggesting that group regularization can partially mitigate the high-dimensionality problem. However, linear projection-based methods such as BPLSDA and BSPLSDA perform poorly on the BRCA multi-class task, yielding Macro-F1 scores of 0.534 and 0.522, respectively, highlighting their limited capacity to model complex nonlinear biological patterns.

\textbf{Performance of Deep Multi-Omics Models.}Recent deep multimodal learning models achieve substantially stronger performance. Methods such as MOGONET effectively capture intra-omics structural relationships using graph convolutional networks, achieving accuracies above 0.80 on ROSMAP and LGG. Similarly, dynamic fusion architectures such as MMDynamic and TMC demonstrate strong performance on BRCA, with MMDynamic reaching an accuracy of 0.861. The contrastive learning framework CLCLSA also performs competitively, achieving an accuracy of 0.825 on ROSMAP. However, these approaches generally rely on the assumption that all omics modalities are available at inference time. Their architectures focus on fusing complete modality inputs rather than actively deciding which modalities should be acquired. As a result, they cannot explicitly account for the acquisition cost associated with different omics assays.

\textbf{Performance of SDM-Q.}
The proposed SDM-Q framework achieves competitive or superior performance across all evaluated datasets while enabling adaptive modality acquisition. On the ROSMAP dataset, SDM-Q achieves the best performance with an accuracy of 0.834 and an AUC of 0.893, outperforming CLCLSA and other baselines. This result indicates that the reinforcement learning--based decision strategy can effectively select informative modalities while suppressing noisy or redundant signals. On the BRCA dataset, SDM-Q achieves an accuracy of 0.861, which is comparable to the best-performing baseline MMDynamic (0.861). Importantly, this performance is achieved while frequently terminating after acquiring only a single modality, demonstrating the efficiency of the proposed cost-aware decision strategy. For the LGG and KIPAN datasets, SDM-Q achieves accuracies of 0.822 and 0.998, respectively, matching the performance of the strongest baselines. These results demonstrate that the proposed framework maintains strong predictive capability across tasks with varying biological complexity.

Overall, SDM-Q achieves competitive classification performance while introducing a cost-aware sequential decision mechanism that dynamically determines the number and order of acquired modalities. By enabling adaptive modality acquisition, the proposed method improves resource efficiency without sacrificing diagnostic accuracy, making it well suited for practical multi-omics analysis scenarios.
\subsection{Ablation Study}

To validate the necessity of the core components within the SDM-Q framework and to elucidate their synergistic mechanisms, we conducted a systematic ablation study. Unlike conventional multimodal research that primarily focuses on feature-fusion gains, this section aims to dissect how three pivotal mechanisms---sequential decision reformulation, cost-aware constraints, and dynamic modality selection---collectively shape the diagnostic strategy after recasting multi-omics classification as a sequential decision-making problem. To this end, we construct five progressive variants, ranging from a static baseline to the complete SDM-Q framework.

(1) \textbf{Baseline:} A standard one-shot classifier that simultaneously uses all modalities without any sequential decision mechanism.

(2) \textbf{B+R:} Introduces RL--based sequential stopping decisions on top of the Baseline, while maintaining a fixed modality acquisition order. In this configuration, the acquisition costs for the second and third modalities are explicitly set to zero, i.e., $C_m=0$, to evaluate the impact of the sequential stopping mechanism independently of cost constraints.

(3) \textbf{B+R+C:} Further incorporates modality-level acquisition cost modeling into B+R by utilizing the specific cost formulations $C_m$ and $\mathrm{Cost}(M_t^{(i)})$ defined in Eq.~\eqref{eq:cost}, enabling the model to explicitly balance diagnostic benefit and acquisition cost.

(4) \textbf{B+R+D:} Introduces dynamic modality selection formulated in  Eq.~\eqref{eq:targetq} on top of B+R, allowing the model to compare candidate modalities through direct action-value (Q-value) estimation based on the current state, but without explicit cost constraints.

(5) \textbf{SDM-Q (B+R+C+D):} The complete framework integrating sequential stopping decisions, modality-level cost modeling, and dynamic modality selection.

The ablation results are summarized in Table \ref{tab:ablation_binary} for binary classification tasks (ROSMAP and LGG) and Table \ref{tab:ablation_multi} for multi-class classification tasks (BRCA and KIPAN).

\begin{table}[H]
\centering
\caption{Ablation results on binary classification tasks (ROSMAP and LGG). The best results are highlighted in bold. $^{*}$ indicates a statistically significant difference ($p<0.05$) compared to our proposed SDM-Q, evaluated by a paired Student t-test.}
\label{tab:ablation_binary}
\small
\resizebox{\textwidth}{!}{%
\begin{tabular}{lcccccc}
\toprule
\multirow{2}{*}{Variant}
& \multicolumn{3}{c}{ROSMAP}
& \multicolumn{3}{c}{LGG} \\
\cmidrule(lr){2-4} \cmidrule(lr){5-7}
& Accuracy & F1-score & AUC
& Accuracy & F1-score & AUC \\
\midrule

Baseline
& 0.761$\pm$0.029$^{*}$
& 0.759$\pm$0.056$^{*}$
& 0.868$\pm$0.019
& 0.764$\pm$0.035
& 0.795$\pm$0.017$^{*}$
& 0.871$\pm$0.006 \\

B+R
& 0.790$\pm$0.032
& 0.800$\pm$0.020$^{*}$
& 0.882$\pm$0.007$^{*}$
& 0.788$\pm$0.019
& 0.796$\pm$0.012$^{*}$
& 0.871$\pm$0.011 \\

B+R+C
& 0.799$\pm$0.016
& 0.809$\pm$0.013$^{*}$
& 0.877$\pm$0.009
& 0.792$\pm$0.029
& 0.793$\pm$0.030$^{*}$
& 0.862$\pm$0.007 \\

B+R+D
& 0.775$\pm$0.023$^{*}$
& 0.786$\pm$0.038$^{*}$
& 0.885$\pm$0.010
& 0.793$\pm$0.027
& 0.806$\pm$0.016$^{*}$
& 0.873$\pm$0.014 \\

\textbf{SDM-Q}
& {\bfseries\boldmath 0.834$\pm$0.025}
& {\bfseries\boldmath 0.834$\pm$0.025}
& {\bfseries\boldmath 0.893$\pm$0.008}
& {\bfseries\boldmath 0.822$\pm$0.011}
& {\bfseries\boldmath 0.822$\pm$0.011}
& {\bfseries\boldmath 0.895$\pm$0.007} \\

\bottomrule
\end{tabular}
}
\end{table}

\begin{table}[H]
\centering
\caption{Ablation results on multi-class classification tasks (BRCA and KIPAN). The best results are highlighted in bold. $^{*}$ indicates a statistically significant difference ($p<0.05$) compared to our proposed SDM-Q model, evaluated by a paired Student t-test.}
\label{tab:ablation_multi}
\small
\resizebox{\textwidth}{!}{%
\begin{tabular}{lcccccc}
\toprule
\multirow{2}{*}{Variant}
& \multicolumn{3}{c}{BRCA}
& \multicolumn{3}{c}{KIPAN} \\
\cmidrule(lr){2-4} \cmidrule(lr){5-7}
& Accuracy & F1-score & Macro-F1
& Accuracy & F1-score & Macro-F1 \\
\midrule

Baseline
& 0.812$\pm$0.019$^{*}$
& 0.812$\pm$0.021$^{*}$
& 0.757$\pm$0.024$^{*}$
& 0.991$\pm$0.003$^{*}$
& 0.991$\pm$0.003$^{*}$
& 0.992$\pm$0.005 \\

B+R
& 0.834$\pm$0.012
& 0.837$\pm$0.013$^{*}$
& 0.793$\pm$0.013
& 0.978$\pm$0.010
& 0.978$\pm$0.010$^{*}$
& 0.962$\pm$0.020 \\

B+R+C
& 0.850$\pm$0.010
& 0.854$\pm$0.010$^{*}$
& 0.817$\pm$0.015
& 0.954$\pm$0.006
& 0.951$\pm$0.007$^{*}$
& 0.910$\pm$0.012 \\

B+R+D
& 0.833$\pm$0.013
& 0.835$\pm$0.017$^{*}$
& 0.792$\pm$0.017
& 0.980$\pm$0.007
& 0.980$\pm$0.007$^{*}$
& 0.972$\pm$0.015 \\

\textbf{SDM-Q}
& {\bfseries\boldmath 0.861$\pm$0.011}
& {\bfseries\boldmath 0.863$\pm$0.013}
& {\bfseries\boldmath 0.828$\pm$0.016}
& {\bfseries\boldmath 0.998$\pm$0.003}
& {\bfseries\boldmath 0.998$\pm$0.003}
& {\bfseries\boldmath 0.999$\pm$0.002} \\

\bottomrule
\end{tabular}
}
\end{table}
\textbf{Baseline and Sequential Stopping Effects.}
Under the Baseline setting, the model performs one-shot multimodal fusion and optimizes purely for classification accuracy. As a result, it cannot account for inter-subject differences in information demand and lacks awareness of modality acquisition cost. Although the baseline model achieves stable performance, it has no mechanism to suppress redundant modality usage, limiting its suitability for resource-constrained clinical scenarios. Introducing sequential stopping in the B+R variant leads to consistent performance improvements. For example, on the ROSMAP dataset, Accuracy and F1-score increase to 0.790 and 0.800, respectively. These results indicate that even under a fixed acquisition order, enabling the model to terminate the diagnostic process adaptively improves predictive robustness. However, the fixed modality order still restricts the model's flexibility and prevents it from adapting to subject-specific modality relevance.

\textbf{Impact of Cost-Aware Modeling.}
The B+R+C variant further introduces modality acquisition cost into the reward function, shifting the optimization objective from purely maximizing classification accuracy to balancing diagnostic performance and acquisition cost. On the ROSMAP dataset, B+R+C maintains a competitive AUC of 0.877 while achieving an Accuracy of 0.799, indicating that the model can regulate modality acquisition without sacrificing discriminative capability. In the BRCA multi-class task, cost-aware modeling also leads to an improved Macro-F1 score. These results suggest that incorporating acquisition cost encourages the model to learn more efficient decision strategies.

\textbf{Role of Dynamic Modality Selection.}
In the B+R+D variant, the model is able to evaluate candidate modalities that have not yet been acquired and dynamically select the most informative one. This mechanism increases the AUC on ROSMAP to 0.885 and slightly improves LGG Accuracy to 0.793. Although the numerical improvements are moderate, they demonstrate that dynamic modality selection enhances information utilization by enabling the model to explore more informative modality combinations.

\textbf{Full SDM-Q.} The complete SDM-Q integrates sequential stopping, cost-aware modeling, and dynamic modality selection, achieving the best or near-best performance across all datasets. For example, on ROSMAP, SDM-Q achieves an Accuracy and F1-score of 0.834, outperforming all ablated variants. On BRCA, the Macro-F1 reaches 0.828, surpassing variants without dynamic decision mechanisms. These results indicate that combining the three components leads to consistently improved predictive performance.

Overall, the ablation results confirm the effectiveness of the proposed decision mechanisms. Sequential stopping enables subject-specific decision granularity, cost-aware modeling regulates modality acquisition under resource constraints, and dynamic selection improves information utilization. Removing any of these components leads to noticeable degradation in the performance--cost trade-off, highlighting the importance of modeling multi-omics diagnosis as a cost-sensitive sequential decision process. These findings further validate the design rationale of SDM-Q and demonstrate the effectiveness of integrating sequential decision-making with cost-aware multi-omics modeling.
\section{Discussion}

In practical clinical scenarios, patients may present with different initial diagnostic information depending on previously available laboratory tests or clinical workflows. Therefore, evaluating the robustness of the proposed SDM-Q framework under different initial modality configurations is important for assessing its practical applicability. In this section, we analyze how the choice of the initial modality influences the final diagnostic performance and explain the rationale behind the configuration adopted in our main experiments.

In the primary experiments reported in Section~\ref{sec:experiments}, mRNA expression was used as the default starting modality during inference. To examine the sensitivity of SDM-Q to this choice, we evaluated three different initial-modality configurations---mRNA expression, DNA methylation, and miRNA expression---across all four benchmark datasets. Under each configuration, the model begins the decision process with the selected modality and subsequently determines whether additional modalities should be acquired according to the learned policy. The resulting classification performance is summarized in Tables \ref{tab:initial_binary} and \ref{tab:initial_multi}.
\begin{table}[H]
\centering
\caption{Effect of different initial modality choices on classification performance for ROSMAP and LGG datasets.}
\label{tab:initial_binary}

{\footnotesize
\renewcommand{\arraystretch}{0.75}
\setlength{\tabcolsep}{8pt}
\begin{tabular}{ccccc}
\hline
Dataset & Initial Omics View & Accuracy & F1-score & AUC \\
\hline

\multirow{3}{*}{ROSMAP~\cite{ref14}} 
& mRNA        & 0.834$\pm$0.025 & 0.834$\pm$0.025 & 0.893$\pm$0.008 \\
& DNAmethyl   & 0.799$\pm$0.025 & 0.798$\pm$0.026 & 0.867$\pm$0.015 \\
& miRNA       & 0.811$\pm$0.030 & 0.811$\pm$0.031 & 0.877$\pm$0.007 \\
\hline

\multirow{3}{*}{LGG~\cite{ref15}} 
& mRNA        & 0.822$\pm$0.011 & 0.822$\pm$0.011 & 0.895$\pm$0.007 \\
& DNAmethyl   & 0.808$\pm$0.028 & 0.806$\pm$0.030 & 0.877$\pm$0.009 \\
& miRNA       & 0.803$\pm$0.021 & 0.802$\pm$0.021 & 0.871$\pm$0.007 \\
\hline

\end{tabular}
}
\end{table}

\begin{table}[H]
\centering
\caption{Effect of different initial modality choices on classification performance for BRCA and KIPAN datasets.}
\label{tab:initial_multi}

{\footnotesize
\renewcommand{\arraystretch}{0.75}
\setlength{\tabcolsep}{8pt}
\begin{tabular}{ccccc}
\hline
Dataset & Initial Omics View & Accuracy & F1-score & AUC \\
\hline

\multirow{3}{*}{BRCA~\cite{ref16}} 
& mRNA        & 0.861$\pm$0.011 & 0.863$\pm$0.013 & 0.828$\pm$0.016 \\
& DNAmethyl   & 0.844$\pm$0.008 & 0.844$\pm$0.009 & 0.817$\pm$0.014 \\
& miRNA       & 0.849$\pm$0.015 & 0.849$\pm$0.016 & 0.819$\pm$0.017 \\
\hline

\multirow{3}{*}{KIPAN~\cite{ref16}} 
& mRNA        & 0.998$\pm$0.003 & 0.998$\pm$0.003 & 0.999$\pm$0.002 \\
& DNAmethyl   & 0.978$\pm$0.005 & 0.978$\pm$0.006 & 0.999$\pm$0.003 \\
& miRNA       & 0.979$\pm$0.007 & 0.979$\pm$0.007 & 0.999$\pm$0.004 \\
\hline

\end{tabular}
}
\end{table}
Overall, the results indicate that mRNA serves as a strong and stable starting modality across datasets. On the ROSMAP and BRCA datasets, initializing the diagnostic process with mRNA consistently leads to the best performance among the three candidates. For example, on ROSMAP, the mRNA-start configuration achieves an Accuracy of 0.834 and an AUC of 0.893, outperforming both the DNA methylation and miRNA configurations. Similar trends are observed on BRCA, where mRNA initialization yields the highest Accuracy and F1-score.

From a biological perspective, this observation is consistent with the role of mRNA expression profiles in reflecting transcriptional activity and downstream functional states of cells. Compared with DNA methylation and miRNA regulation signals, mRNA expression often captures more direct phenotypic manifestations of disease-related molecular processes~\cite{ref31,ref32}. As a result, mRNA data frequently contain strong discriminative signals for disease classification, making them an informative starting point for sequential decision-making.

Based on both the empirical observations and biological considerations, mRNA was selected as the default initial modality in the main experiments. This configuration enables the model to begin the decision process with a highly informative modality while allowing additional modalities to be acquired only when necessary. Consequently, the SDM-Q framework can achieve a favorable balance between diagnostic accuracy and information acquisition efficiency. These findings suggest that selecting an informative initial modality can improve the efficiency of sequential diagnostic strategies, further highlighting the importance of modality-aware decision modeling in multi-omics precision medicine.

Despite the promising results, several limitations remain. The experimental evaluation in this study is conducted primarily on publicly available multi-omics benchmark datasets, where all modalities are pre-collected. Although modality acquisition costs are simulated through predefined cost settings, real-world clinical workflows often involve additional complexities such as heterogeneous testing procedures, varying turnaround times, and cross-institutional variability. These factors may influence the practical deployment of sequential diagnostic strategies. Future work will focus on validating the proposed framework in more realistic clinical environments and exploring more interpretable decision mechanisms. In particular, incorporating clinical expert knowledge into the reward design and developing more transparent modality-selection strategies may further enhance the clinical applicability of SDM-Q.
\section{Conclusion}

This study proposes SDM-Q, a deep Q-learning-based framework for dynamic multi-omics diagnosis with on-demand modality acquisition. Unlike conventional static approaches that require all omics modalities to be collected prior to prediction, SDM-Q reformulates the multi-omics diagnostic process as a finite-horizon sequential decision-making problem. At each decision stage, the framework evaluates whether to acquire additional modalities or terminate the diagnostic process based on the currently observed omics information. By explicitly incorporating modality acquisition costs into the reward function, SDM-Q enables adaptive, subject-specific decisions regarding both the number and the order of acquired modalities, thereby identifying efficient diagnostic pathways for individual patients.

Extensive experiments on four public multi-omics benchmark datasets demonstrate that SDM-Q can substantially reduce redundant modality acquisition while maintaining competitive and stable diagnostic performance. The results indicate that, for a considerable proportion of subjects, accurate predictions can be achieved without requiring complete multi-omics profiles. These findings highlight the effectiveness of modeling multi-omics diagnosis as a cost-aware sequential decision problem and demonstrate the potential of adaptive modality acquisition strategies for improving the efficiency of multi-omics data utilization. Overall, SDM-Q provides a flexible and cost-efficient framework for dynamic multi-omics diagnosis and offers a promising direction toward resource-aware precision medicine.

\section*{Acknowledgment}

Nan Mu was supported by the National Natural Science Foundation of China (grant number 62006165) and the Natural Science Foundation of Sichuan Province (grant number 2025ZNSFSC1477).



\bibliographystyle{elsarticle-num}
\bibliography{cas-refs}
\end{document}